\documentclass[letterpaper, 10 pt, conference]{ieeeconf}  

\IEEEoverridecommandlockouts                              

\usepackage{caption}
\usepackage{subcaption}

\usepackage{amsmath}
\usepackage{amssymb}
\usepackage{graphicx}
\usepackage{tikz}
\usepackage{pgfplots}
\pgfplotsset{compat=1.18}
\usepgfplotslibrary{groupplots}
\usepgfplotslibrary{fillbetween}
\usepackage[
  backend=biber,
  style=ieee,
  giveninits=true,
  maxnames=2,
  minnames=1
]{biblatex}
\AtBeginDocument{\let\cite\parencite}
\usepackage{mathtools}
\mathtoolsset{showonlyrefs} 

\usepackage{acronym}
\acrodef{PPO}{Proximal Policy Optimization}
\acrodef{RL}{Reinforcement Learning}
\acrodef{BC}{Behavioral cloning}
\acrodef{AP}{actor-only pretraining}
\acrodef{ACP}{actor-critic pretraining}
\acrodef{NP}{no pretraining}
\acrodef{PIRL}{pretraining with imitation and RL fine-tuning}

\usepackage{xcolor}
\definecolor{myorange}{RGB}{245, 130, 32}
\definecolor{mygreen}{RGB}{23, 156, 125}
\definecolor{myblue}{RGB}{0, 91, 127}
\definecolor{mygrey}{RGB}{166, 187, 200}
\definecolor{myteal}{RGB}{0, 133, 152}
\definecolor{myturquoise}{RGB}{57, 193, 205}
\definecolor{myyellow}{RGB}{253, 185, 19}
\definecolor{mypink}{RGB}{187, 0, 86}

\title{\LARGE \bf
Actor-Critic Pretraining for Proximal Policy Optimization
}

\author{Andreas~Kernbach$^{1,2}$,
        Amr~Elsheikh$^{2}$,
        Nicolas~Grupp$^{3}$,
        René Nagel$^{4}$,
        and~Marco~F.~Huber$^{1,2}$
\thanks{$^{1}$Fraunhofer Institute for Manufacturing Engineering and Automation IPA, Stuttgart, Germany}%
\thanks{$^{2}$Institute of Industrial Manufacturing and Management IFF, University of Stuttgart, Germany}%
\thanks{$^{3}$Institute for Control Engineering of Machine Tools and Manufacturing Units ISW, University of Stuttgart, Germany}%
\thanks{$^{4}$Materials Testing Institute MPA, University of Stuttgart, Germany}%
}

\begin{document}

\maketitle
\thispagestyle{empty}
\pagestyle{empty}

\begin{abstract}

Reinforcement learning (RL) actor–critic algorithms enable autonomous learning but often require a large number of environment interactions, which limits their applicability in robotics.
Leveraging expert data can reduce the number of required environment interactions. A common approach is actor pretraining, where the actor network is initialized via behavioral cloning on expert demonstrations and subsequently fine-tuned with RL.
In contrast, the initialization of the critic network has received little attention, despite its central role in policy optimization. This paper proposes a pretraining approach for actor–critic algorithms like Proximal Policy Optimization (PPO) that uses expert demonstrations to initialize both networks. The actor is pretrained via behavioral cloning, while the critic is pretrained using returns obtained from rollouts of the pretrained policy. The approach is evaluated on 15 simulated robotic manipulation and locomotion tasks. Experimental results show that actor–critic pretraining improves sample efficiency by 86.1\% on average compared to no pretraining and by 30.9\% to actor-only pretraining.

\end{abstract}

\section{Introduction}
\ac{RL} is an artificial intelligence paradigm in which an agent learns a policy through direct interaction with an environment. By observing states, taking actions, and receiving rewards, the agent aims to maximize discounted cumulative reward over time.
Among the most prominent \ac{RL} algorithms are actor–critic methods, like \ac{PPO}. There, the actor selects actions according to a parameterized policy, while the critic estimates the value of a given state, providing feedback to guide policy updates.
In practice, both the actor and the critic are implemented using neural networks. These networks are typically initialized with random parameters and are iteratively refined during training.
Despite their success, \ac{RL} methods are often criticized for their sample inefficiency. The reward signal usually carries limited information as it is one scalar value and may be delayed. In conjunction with the need for exploration and a permanent distribution shift, \ac{RL} requires a high number of interactions with the environment.

Imitation learning and in particular \ac{BC} have been proposed to mitigate this limitation. By pretraining the actor network on expert demonstrations, the \ac{RL} algorithm does not start from random initialization but from a policy that already imitates expert behavior, thereby reducing the number of samples required during subsequent \ac{RL} fine-tuning.
However, existing approaches typically focus on initializing only the actor network, while initialization strategies for the critic network have received less attention.
This work addresses the research gap by providing an approach for critic initialization that leads to increased sample efficiency and improved convergence.
We focus on \ac{PPO} as one representative and widely used actor-critic algorithm. The pretraining concepts can also be transferred to other actor-critic algorithms, but require some algorithmic dependent adjustments.
The core contributions are:

\begin{itemize}
    \item A theoretical pretraining approach for actor and
    critic networks adjusted to \ac{PPO}.
    \item An empirical evaluation demonstrating improved sample efficiency and
    algorithm convergence across 15 benchmark environments.
\end{itemize}

\begin{figure}
    \centering
    \includegraphics[width=1\linewidth]{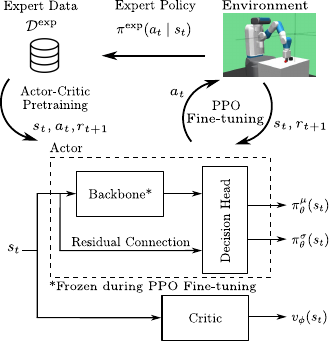}
    \caption{Visualization of the pretraining and fine-tuning approach using expert data with an actor-critic model. First an expert policy is used to pretrain actor and critic, which are subsequently fine-tuned using PPO.}
    \label{fig:ch1_concept_image}
    \vspace*{-0.6cm}
\end{figure}
\section{Related Work}
\label{sec:related_work}

Current \ac{RL} is criticized for sample inefficiency. Learning relies on trial-and-error interaction, where reward signals are scalar, may be sparse or delayed, and where a lot of exploration is needed, with constant distribution shifts in actions and observations. As a result, a large number of environment interactions is required to obtain reliable policies~\cite{Correia2024}.
This issue is particularly pronounced in on-policy algorithms, like the widely used \ac{PPO}~\cite{Schulman2017}, which discard collected data after each policy update and therefore cannot reuse past experience, further increasing sample requirements.
In real-world robotic applications, interactions require time and lead to physical wear on the hardware. Moreover, the \ac{RL} training can produce unsafe actions, such as collisions.

Imitation Learning has been proposed as a complement to \ac{RL} to increase sample efficiency and reduce safety concerns by learning from expert demonstrations. 
A common Imitation Learning approach is \ac{BC}, which trains a policy via supervised learning on expert state–action pairs. \ac{BC} has been successfully applied in robotic manipulation tasks such as object placement, tool hanging or robotic connector installation~\cite{mandlekar22a, Kernbach2026}. However, it suffers from compounding errors, as the learned policy can make small mistakes and deviate into states not represented in the demonstration data~\cite{Zhao2023}.
One strategy to combine the strengths of \ac{BC} and \ac{RL} is to pretrain the policy using \ac{BC} on expert data and subsequently fine-tune it with \ac{RL}. The \ac{BC} policy initialization reduces the number of samples \ac{RL} requires for exploration, mitigates safety risks during training, but allows \ac{RL} to adjust the policy to reduce compounding errors. At the same time, it reduces the distribution shift during \ac{RL} training, since the policy starts closer to an optimal policy. Such pretraining has been shown to reduce the sample requirements for dexterous manipulation tasks ~\cite{Rajeswaran2017, Wagenmaker2025} and has also been applied with \ac{PPO} to robotic assembly tasks~\cite{Ankile2025}.

Most prior work focuses on initializing the actor network, as it directly determines action selection. However, comparatively little attention is given to effective critic initialization. One related approach is \ac{PIRL}, which pretrains the actor using \ac{BC} and subsequently fine-tunes with \ac{PPO} while keeping the actor weights frozen, allowing only the critic to be updated. After the critic loss converges, the actor is unfrozen and both networks are jointly optimized~\cite{Ramrakhya2023}. Although being a step in a promising direction, this heuristic can be unstable, evident by the initial drop in performance and the delayed improvement beyond the expert baseline observed in the paper’s own experimental results. Another \ac{PPO} initialization approach initializes the critic to produce value estimates centered around a predefined value improving training stability and performance~\cite{Jang2023}. Other alternatives include auxiliary losses that regularize the policy toward the expert during pretraining~\cite{Schmitt2018}, and, for off-policy methods, warm-up phases seeded with rollouts from a pretrained policy~\cite{ICLR2025_50449129}.

However, as this work focuses on \ac{PPO}, a research gap remains in finding alternative and more general strategies for critic initialization in \ac{PPO} and other actor–critic \ac{RL} algorithms.

\section{Problem Formulation}
\label{sec:problem_formulation}

We consider a Markov Decision Process (MDP)
\linebreak 
$\mathcal{M} = (\mathcal{S}, \mathcal{A}, P, R, \gamma)$,
where $\mathcal{S}$ denotes the state space, $\mathcal{A}$ the action space,
$P$ the transition dynamics,
$R:\mathcal{S}\times\mathcal{A}\rightarrow\mathbb{R}$ the reward function,
and $\gamma \in (0,1]$ the discount factor.
Each episode $i$ has a finite horizon $T^i \in \mathbb{N}$, with environment steps indexed by
$t = 0, \dots, T^i - 1$, where $T^i$ may vary due to early termination or truncation.
The discounted return starting at environment step $t$ until an episode end is defined as
\begin{equation}
\label{eq:return}
G_t = \sum_{k=t}^{T-1} \gamma^{k-t} r_{k+1}~.
\end{equation}
We denote by $G_{t=0}$ the episodic return starting from the initial step $t=0$ until the end of the episode.
At each environment step $t$, the policy $\pi(a_t \mid s_t)$ represents the
probability of selecting action $a_t$ given state $s_t$. Formally, a stochastic policy is
defined as
$
\pi(a_t \mid s_t) \colon \mathcal{S} \rightarrow \mathcal{P}(\mathcal{A}),
$
where $\mathcal{P}(\mathcal{A})$ denotes the space of probability distributions
over the action space $\mathcal{A}$.
The state-value function under policy $\pi$ is defined as
\begin{equation}
\label{eq:value_function}
V^{\pi}(s_t)
= \mathbb{E}_{\pi}\!\left[ G_t \mid s_t \right]~,
\end{equation}
i.e., the expected discounted return when starting from state $s_t$ and
following policy $\pi$ thereafter.
We assume access to an non-optimal but reasonably good expert policy $\pi^{\mathrm{exp}}(a_t \mid s_t)$ that achieves an undiscounted episodic return $G_{t=0}^{\mathrm{exp}}$ on average.
The expert policy can be executed in the environment for a total of
$n^{\mathrm{exp}}$ environment steps, distributed across
$N^{\mathrm{exp}}$ episodes.
Each episode $i$ has a finite horizon $T^{i}$, which produces an offline dataset 
\begin{equation}
\mathcal{D}^{\mathrm{exp}}
= \bigcup_{i=1}^{N^{\mathrm{exp}}}
\{(s^{i}_t, a^{i}_t, r^{i}_{t+1})\}_{t=0}^{T^{i}-1}~,
\end{equation}
of environment transitions $s_t \rightarrow a_t \rightarrow r_{t+1}$ where $\sum_{i=1}^{N^{\mathrm{exp}}} T^{i} = n^{\mathrm{exp}}$.

In \ac{PPO} two parameterized neural
networks are employed: an actor network representing the policy
$\pi_{\theta}(a_t \mid s_t)$ and a critic network $v_{\phi}(s_t)$ that approximates the
state-value function $V^{\pi}(s_t)$.
In conventional \ac{PPO} training, no pretraining is performed and both the actor parameters $\theta$ and the critic
parameters $\phi$ are initialized randomly. Both networks are then
optimized through on-policy interaction with the environment. The training can run for an arbitrary amount of environment steps, but in this work, training is considered complete once the average episodic return surpasses the expert-level performance and reaches an environment specific and user-defined target return $G_{t=0}^{\mathrm{tar}}$ averaged over sufficiently many episodes.
We denote by $n_{\mathrm{PPO}}^{\mathrm{NP}}$ the number of environment
steps required by \ac{PPO} with no pretraining to reach this
return level.

We assume that performing environment steps is expensive, as is the case in robotics.
Therefore, we aim for high sample efficiency to minimize the total number of environment steps.
We seek an \ac{ACP} and fine-tuning procedure that reduces the total number of
required environment steps compared to \ac{PPO} with \ac{NP} and \ac{AP}.
We denote by $n^{\mathrm{tot}}$ the total number of environment steps required for training, including expert data collection, rollout and
subsequent \ac{PPO} fine-tuning.

Formally, we want a sample efficient approach that achieves a considerable reduction in environment steps so that
$n^{\mathrm{tot}}_{\mathrm{ACP}}
\ll n_{\mathrm{PPO}}^{\mathrm{NP}}
\quad \text{and} \quad
n^{\mathrm{tot}}_{\mathrm{ACP}}
\ll n_{\mathrm{AP}}^{\mathrm{tot}},$
averaged over different environments.

\section{Methodology}
The methodology explains the core method of actor and critic pretraining and the extended step limit and residual model architecture features.
\subsection{Actor Pretraining}
\label{sec:actor_pretraining}

We assume a Gaussian policy for the actor network 
$\pi_{\theta}(a_t \mid s_t)$ with mean $\pi_{\theta}^{\mu}(s_t)$ and diagonal
covariance $\pi_{\theta}^{\sigma}(s_t) \mathbf I$, where $\mathbf I$ denotes the identity matrix, i.e.,
\begin{equation}
\pi_{\theta}(a_t \mid s_t)
= \mathcal{N}\!\left(a_t \mid \pi_{\theta}^{\mu}(s_t), \pi_{\theta}^{\sigma}(s_t) \mathbf I\right)~,
\end{equation}
and $\pi_{\theta}^{\sigma}(s_t)=e^{-2}$ is fixed during pretraining, with $e$ denoting Euler's number. This value is taken from the default hyperparameter settings of the RL Baselines3 Zoo library~\cite{rl-zoo3}. The chosen distribution assumption implies a continuous action space $\mathcal{A}$. For a discrete action space $\pi_{\theta}(a_t \mid s_t)$ can be parameterized differently without affecting the pretraining procedure.

The actor is pretrained using \ac{BC} on the expert dataset
$\mathcal{D}^{\mathrm{exp}}$ by minimizing the action space dependent norm between
expert actions and predicted mean actions, which corresponds to solving
\begin{equation}
\theta^{\ast}
= \arg\min_{\theta}
\left\lVert a_t - \pi_{\theta}^{\mu}(s_t) \right\rVert^2_{\mathcal{A}}.
\end{equation}
over all steps $t$ in $\mathcal{D}^{\mathrm{exp}}$. In a continuous action space this corresponds to minimizing the mean-squared error.

\subsection{Critic Pretraining}
The critic represents the value function under a given
policy, as stated in Eq.~\eqref{eq:value_function}. But since the pretrained policy $\pi_{\theta}(a_t \mid s_t)$ only approximates the expert policy, the returns of that policy may not match up with the returns obtained from $\mathcal{D}^{\mathrm{exp}}$. We therefore hypothesize that additional rollouts are required to obtain critic training targets that are consistent with the evaluated policy. Accordingly, we perform rollouts using the pretrained policy $\pi_{\theta}(a_t \mid s_t)$.
This results in an additional rollout dataset
\begin{equation}
\mathcal{D}^{\mathrm{rol}}
= \bigcup_{i=1}^{N^{\mathrm{rol}}}
\{(s^{i}_t, a^{i}_t, r^{i}_{t+1})\}_{t=0}^{T^{i}-1}~.
\end{equation}
For each transition in $\mathcal{D}^{\mathrm{rol}}$, returns $G_{t}^{\mathrm{rol}}$ are computed as stated in Eq.~\eqref{eq:return}.

\ac{PPO} maximizes a composite loss function consisting of
a clipped policy loss $\mathcal{L}^{\mathrm{CLIP}}_t(\theta)$, a value loss $\mathcal{L}^{\mathrm{VF}}_t(\phi)$, and an entropy regularization term $S\!\left[\pi_{\theta}\right](s_t)$ resulting in
$\mathcal{L}_{t}^{\mathrm{PPO}}(\theta, \phi)
= 
\mathcal{L}^{\mathrm{CLIP}}_t(\theta)
- c_1 \mathcal{L}^{\mathrm{VF}}_t(\phi)
+ c_2 S\!\left[\pi_{\theta}\right](s_t)
~,$
averaged over all steps $t$ in an on-policy rollout batch, with 
$c_1, c_2$ being hyperparameters.
The value loss is defined as
\begin{equation}
\label{eq:value_loss}
\mathcal{L}^{\mathrm{VF}}_t(\phi)
= \left( v_{\phi}(s_t) - V_{t}^{\mathrm{tar}} \right)^2,
\end{equation}
with the output of the critic network $v_{\phi}(s_t)$ that is trained to approximate the state-value
function of the current policy. The target value $V_{t}^{\mathrm{tar}}$ is treated as constant with respect to $\phi$.
During \ac{PPO} training, target values are computed using the generalized advantage estimate
$\hat{A}_t
= \sum_{l=0}^{T-t-1} (\gamma \lambda)^l \delta_{t+l}~, $
with a decay parameter $\lambda \in [0,1]$ and temporal-difference residual
$\delta_t
= r_t + \gamma v_{\phi_{\mathrm{old}}}(s_{t+1}) - v_{\phi_{\mathrm{old}}}(s_t)~,$
where $v_{\phi_{\mathrm{old}}}$ denotes the critic model that is treated as constant with respect to $\phi$.
The corresponding value targets are computed as
$V_{t}^{\mathrm{tar}}
= \hat{A}_t + v_{\phi_{\mathrm{old}}}(s_t)~.$

In \ac{PPO}, the advantage function quantifies the deviation of the observed return
from the value predicted by the critic. Under a perfect value prediction, the advantage is expected to be zero on average,
$\mathbb{E}_{\pi}[\hat A_t \mid s_t] = 0$, and the critic
$v_{\phi}(s_t)$ predicts $\mathbb{E}_{\pi}[ G_t \mid s_t ]$.
For critic pretraining, we assume perfect prediction in expectation. The rollout return $G^{\mathrm{rol}}_t$ is generated by following the same policy
$\pi_{\theta}(a_t \mid s_t)$ that the critic is intended to approximate. Thus, we set the target value equal to the observed rollout return, i.e.,
$V^{\mathrm{tar}}_t = G^{\mathrm{rol}}_t~.$

In line with the value loss in Eq.\eqref{eq:value_loss} the critic is pretrained by minimizing the mean squared error
between the predicted value and the rollout return. Formally, the critic
parameters are obtained by solving
\begin{equation}
\phi^{\ast}
= \arg\min_{\phi}\;
\left( v_{\phi}(s_t) - G^{\mathrm{rol}}_t \right)^2
\end{equation}
over all steps $t$ in $\mathcal{D}^{\mathrm{rol}}$.

\subsection{Extended Step Limit}

As stated in Sec.~\ref{sec:problem_formulation}, all considered environments have a finite horizon $T$. However, in some environments this horizon is introduced artificially by truncating episodes after a fixed number of steps. Without this truncation, episodes may run longer until a terminal state is reached, or may not terminate at all. In these untruncated cases where $T\rightarrow \infty$, the return can be higher or even form an infinite series, denoted by $G_t^{\infty}$.
This artificial truncation introduces a bias in value estimation, as the remaining tail of the potential return is ignored. 
To address this issue, we introduce an \emph{extended step limit}. Due to discounting ($\gamma < 1$), distant rewards contribute negligibly.
We exploit this property by imposing an extended step limit
$T^{\mathrm{ext}} \geq T$ on rollout episodes, where $T$ denotes the nominal
episode horizon.
The goal is to choose $T^{\mathrm{ext}}$ such that the truncation error $\epsilon_t := G_t^{\infty} - G^{\mathrm{rol}}_t$ satisfies $|\epsilon_t| \leq \tau$ for all $t < T$, where $\tau > 0$ is a tolerance chosen based on the scale of returns in each environment.
Assuming rewards are bounded by an upper bound $r_{\max}$, i.e.,
$|r_t| \leq r_{\max}$ for all $t$, the extended step limit guaranteeing $|\epsilon_t| \leq \tau$ for all $t < T$ is given by
\begin{equation}
T^{\mathrm{ext}} =  \left\lceil T + \frac{\ln\left(\frac{\tau(1-\gamma)}{r_{\max}}\right)}{\ln(\gamma)} \right\rceil,
\label{eq:extended_step_limit}
\end{equation}
with the restriction $\tau \le r_{\mathrm{max}}/(1-\gamma)$.
The derivation bounds the truncation error using a geometric series argument: truncating at step $T^{\mathrm{ext}}$ leaves a tail bounded by
$r_{\max} \cdot \gamma^{T^{\mathrm{ext}}-T}/(1-\gamma)$, and solving for $T^{\mathrm{ext}}$ such that this bound equals $\tau$ yields Eq.~\eqref{eq:extended_step_limit}. Since errors propagate
backward scaled by $\gamma < 1$, ensuring $|\epsilon_{T}| \leq \tau$ guarantees $|\epsilon_t| < \tau$ for all earlier steps.

\subsection{Residual Model Architecture}
We propose a specific model architecture as a complimenting feature for our pretraining approach.
The actor network $\pi_{\theta}(a_t \mid s_t)$ consists of a
backbone network and a decision head, connected via a residual connection, as
illustrated in Fig.~\ref{fig:ch1_concept_image}. The backbone processes the observation $s_t$ and extracts a latent
feature representation. These latent features are then passed to the decision head together with the original observation through a residual connection.
The decision head combines both inputs to predict the final action
distribution. During the pretraining phase, all actor parameters $\theta$ are optimized.
During subsequent \ac{PPO} fine-tuning, the parameters of the backbone network are
kept frozen, while only the parameters of the decision head are updated. The residual connection provides the decision head with direct access to the original observation $s_t$, so fine-tuning can still condition actions on the original state even if the backbone features are imperfect.
This architectural design aims to preserve the expert-induced behavior learned during pretraining, while not restricting policy updates during \ac{PPO} fine-tuning, similar to~\cite{Ramrakhya2023}. Formulated more conceptually, the actor is designed to retain an immutable expert instinct that guides decision making while still being flexible for learning. Feedforward networks with ReLU activations are used for the actor and critic. The number of layers and neurons per layer are chosen based on the default network hyperparameters from RL Baselines3 Zoo library.

\section{Evaluation}

The proposed pretraining approach is evaluated on 15 simulated benchmark
environments covering robotic manipulation and locomotion tasks from the Gymnasium and Gymnasium-Robotics benchmark suite~\cite{gymnasium, gymnasium_robotics}.
We compare four pretraining and fine-tuning methods: (1) actor-critic pretraining \ac{ACP}, (2) actor-only pretraining \ac{AP}, and (3) no pretraining \ac{NP}, all with subsequent \ac{PPO} fine-tuning. In addition, we include (4) an alternative state-of-the-art pretraining approach PIRL (see Sec.~\ref{sec:related_work}) for comparison.

\subsection{Sample Efficiency Evaluation}
\label{sec:sample_efficiency_evaluation}

As stated in Section~\ref{sec:problem_formulation}, we assume access to an expert
policy.
Such a policy may originate from a human expert, a heuristic controller, or another
previously trained \ac{RL} agent.
For a proof of concept of our approach and convenience, we use expert policies
$\pi^{\mathrm{exp}}(a_t \mid s_t)$ obtained from the RL Baselines3 Zoo
library.
For each environment, the library provides recommended \ac{RL} algorithms with tuned
hyperparameters, as well as empirically achievable undiscounted ($\gamma=1$) target returns
$G_{t=0}^{\mathrm{tar}}$.
We train the corresponding expert policy using the recommended configuration
and terminate \ac{RL} training early once an expert return
$G_{t=0}^{\mathrm{exp}}$ is reached. 
The expert return is intentionally set below the target return, i.e.,
$
G_{t=0}^{\mathrm{exp}} = c \cdot G_{t=0}^{\mathrm{tar}}, \quad 0 < c < 1,
$
ensuring that further performance gains through \ac{PPO} fine-tuning are possible.
In this work, $c = 0.65$ is used, as \ac{AP} was empirically observed to
converge to approximately this fraction of the target return in one environment (see \texttt{Reacher} in Fig.~\ref{fig:ch5_n_expert_evaluation}). Thus, in all subsequent experiments the number of expert demonstration steps $n^{\mathrm{exp}}$ is fixed for all pretraining approaches.

\begin{figure}[t]
    \centering
    \vspace*{0.17cm}
    \input{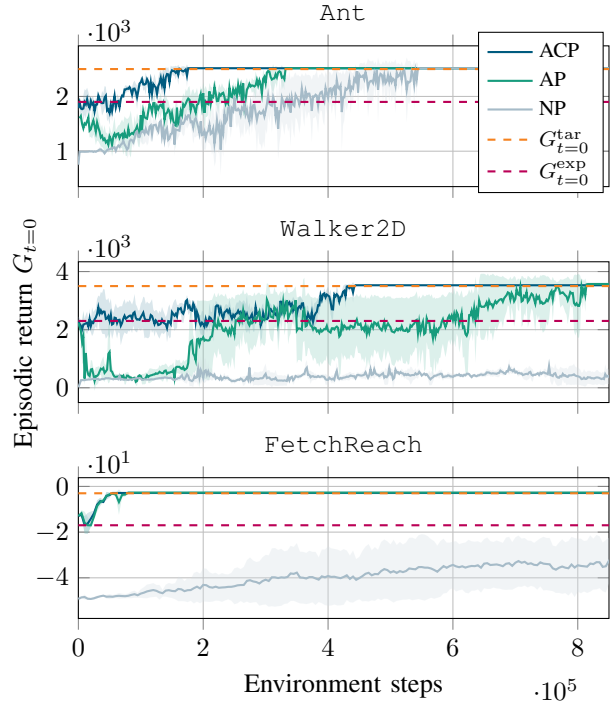} 
    \caption{Comparison of the learning results in \ac{PPO} fine-tuning given \ac{ACP}, \ac{AP} and \ac{NP} in three exemplary environments. Returns are averaged over three random seeds for fine-tuning, with shaded regions
    indicating the standard deviation.}
    \label{fig:ch5_ACP_AP_PPO_return_plot}
    \vspace*{-0.6cm}
\end{figure}

The hyperparameters (e.g., learning rate, number of rollouts) of all stated results (both pretraining and PPO fine-tuning in all Figures and Table) are optimized using the Optuna framework~\cite{optuna_2019}.
All reported returns are obtained by averaging the undiscounted episodic return over 30 evaluation episodes. 
The training budget for \ac{PPO} fine-tuning is set to $10^{6}$ environment steps for all environments, following the configuration used in the RL Baselines3 Zoo library.

The learning curves of \ac{PPO} fine-tuning for three exemplary environments are shown in Fig.~\ref{fig:ch5_ACP_AP_PPO_return_plot}, where the benefit of pretraining is evident. The pretraining results converge faster than those without pretraining. Moreover, in the \texttt{Walker2D} and \texttt{FetchReach} environments, \ac{PPO} without pretraining
fails to reach the target return $G_{t=0}^{\mathrm{tar}}$ within the given
training budget. 
Catastrophic forgetting is a known issue in \ac{AP}.
This behavior is visible in the \texttt{Ant} and \texttt{Walker2D} environments, where the return
drops below expert-level performance during the first $2 \cdot 10^{5}$ environment
steps.
In contrast, \ac{ACP} mitigates catastrophic forgetting in some evaluated environments.

To summarize the results across all environments,
Tab.~\ref{tab:sample_efficiency} reports the number of environment 
steps required to reach the target return for all four pretraining methods. In the cases where \ac{NP} converged to target-level returns, \ac{AP} reduces the required number of environment steps
by 69.0\% 
compared to \ac{NP}, while \ac{ACP} achieves a reduction of 86.1\% 
relative to \ac{NP} on average. 
Thus, \ac{ACP} increases the sample efficiency compared to \ac{AP} by 30.9\% 
on average.
These results empirically show that pretraining improves sample efficiency on average and
that initializing the critic in addition to the actor provides further gains
beyond state-of-the-art \ac{BC}.

In addition, \ac{ACP} is compared to \ac{PIRL}, where the actor weights are kept frozen during \ac{PPO} training and only the critic is updated. Once the value loss (see Eq.~\eqref{eq:value_loss}) converges, the actor weights are unfrozen. We define critic convergence as a 10\% change rate in the value loss and denote the required number of steps with frozen actor by $n^{\mathrm{fro}}_{\mathrm{PPO}}$. In 13 out of 15 environments (86.6\%) \ac{ACP} requires less environment steps with an average sample reduction of 20.5\%.

However not all environments benefited from critic pretraining more than from \ac{AP}. In three of the 15 environments (20.0\%), neither \ac{ACP} nor \ac{PIRL} resulted in reduced sample efficiency compared to \ac{AP}. Although the observed degradation was at most -9.2\% (see \texttt{InvertedDoublePendulum} in Tab.~\ref{tab:sample_efficiency}), this outcome suggests that the usefulness of critic pretraining may be environment dependent and that further investigation is required to understand when and why it may not be beneficial. A common property of the \texttt{Humanoid} family, that does not profit from critic pretraining is their substantially higher observation space dimensionality compared to the other environments.

\newcommand{\hcell}[1]{\raisebox{0pt}[2.8ex][1.5ex]{#1}}
\begin{table*}[t]
  \centering
  \vspace*{0.17cm} 
  \setlength{\tabcolsep}{4pt}
  \begin{tabular}{c c p{1pt} c c c p{1pt} c c c p{1pt} c c c p{1pt} c c c}
    \hline
    Environment
    & \multicolumn{1}{c}{\hcell{NP}}
    &
    & \multicolumn{1}{c}{\hcell{}}         
    & \multicolumn{2}{c}{\hcell{AP}}
    &
    & \multicolumn{3}{c}{\hcell{PIRL}}
    &
    & \multicolumn{3}{c}{\hcell{ACP}}
    &
    & \multicolumn{3}{c}{\hcell{Sample Reduction}} \\
    \cline{2-2}\cline{5-6}\cline{8-10}\cline{12-14}\cline{16-18}
    & \hcell{$n^{\mathrm{NP}}_{\mathrm{PPO}}$}
    &
    & \hcell{$n^{\mathrm{exp}}$}
    & \hcell{$n^{\mathrm{fine}}_{\mathrm{PPO}}$}
    & \hcell{$n^{\mathrm{tot}}_{\mathrm{AP}}$}
    &
    & \hcell{$n^{\mathrm{fro}}_{\mathrm{PPO}}$}
    & \hcell{$n^{\mathrm{fine}}_{\mathrm{PPO}}$}
    & \hcell{$n^{\mathrm{tot}}_{\mathrm{PIRL}}$}
    &
    & \hcell{$n^{\mathrm{rol}}$}
    & \hcell{$n^{\mathrm{fine}}_{\mathrm{PPO}}$}
    & \hcell{$n^{\mathrm{tot}}_{\mathrm{ACP}}$}
    &
    & \hcell{NP}
    & \hcell{AP}
    & \hcell{PIRL} \\
    \hline
    \texttt{Walker2D}          & $-$   & & 0.6 & 190.0 & 190.6 & & 12.2 & 55.8 & 68.6 & & 49.5 & 6.0 & \textbf{56.1} & & 100\% & 70.6\% & 18.2\% \\
    \texttt{Hopper}            & 588.0 & & 1.0 & 86.0  & 87.0  & & 11.8 & 76.2 & 89.0 & & 19.9 & 10.0 & \textbf{30.9} & & 94.7\% & 64.5\% & 65.3\% \\
    \texttt{Ant}               & 544.0 & & 1.0 & 190.0 & 191.0 & & 55.3 & 100.7 & 157.0 & & 1.7  & 96.0 & \textbf{98.7} & & 81.9\% & 48.3\% & 37.1\% \\
    \texttt{Humanoid}          & $-$   & & 0.1 & 52.0  & \textbf{52.1} & & 45.6 & 42.4 & 88.1 & & 1.0 & 52.0 & 53.1 & & 100\% & -1.9\% & 39.7\% \\
    \texttt{InvertedD.Pendulum}& 115.0 & & 10.4& 0.5   & \textbf{10.9} & & 1.7 & 0.3 & 12.4 & & 1.0 & 0.5  & 11.9 & & 89.7\% & -9.2\% & 4.0\% \\
    \texttt{InvertedPendulum}  & 27.4  & & 0.1 & 4.4   & 4.5   & & 0.4 & 3.9 & 4.4 & & 1.0 & 1.9  & \textbf{3.1} & & 88.7\% & 31.5\% & 29.5\%\\
    \texttt{Pusher}            & $-$   & & 0.1 & 124.0 & 124.1 & & 20.7 & 67.3 & 88.1 & & 9.7 & 64.0 & \textbf{73.8} & & 100\% & 40.5\% & 16,3\%\\
    \texttt{Reacher}           & 272.0 & & 0.8 & 196.0 & 196.8 & & 18.4 & 17.6 & \textbf{36.8} & & 19.9 & 38.0 & 58.7 & & 78.4\% & 70.2\% & -59.5\%\\
    \texttt{HalfCheetah}       & $-$   & & 1.0 & 72.0  & 73.0  & & 26.1 & 61.9 & 89.0 & & 1.6 & 58.0 & \textbf{60.6} & & 100\% & 17.0\% & 31.9\%\\
    \texttt{Swimmer}           & $-$   & & 1.0 & 54.0  & 55.0  & & 34.8 & 57.2 & 93.0 & & 2.2 & 46.0 & \textbf{49.2} & & 100\% & 10.5\% & 47.1\%\\
    \texttt{HumanoidStandup}   & 405.0 & & 1.0 & 65.0  & \textbf{66.0} & & 27.6 & 62.4 & 91.0 & & 2.6 & 65.0 & 68.6 & & 83.1\% & -3.9\% & 24.6\%\\
    \texttt{FetchPush}         & $-$   & & 35.0& 140.0 & 175.0 & & 24.6 & 55.4 & 115.0 & & 1.0 & 75.0 & \textbf{111.0} & & 100\% & 36.6\% & 3.5\%\\
    \texttt{FetchPickAndPlace} & $-$   & & 30.0& 180.0 & 210.0 & & 14.3 & 45.7 & \textbf{90.0} & & 1.0 & 75.0 & 106.0 & & 100\% & 49.5\% & -17.8\%\\
    \texttt{FetchSlide}        & $-$   & & 50.0& 60.0  & 110.0 & & 53.9 & 41.1 & 145.0 & & 1.0 & 40.0 & \textbf{91.0} & & 100\% & 17.3\% & 37.2\%\\
    \texttt{FetchReach}        & $-$   & & 0.5 & 70.0  & 70.5  & & 14.3 & 63.7 & 78.5 & & 10.0& 44.0 & \textbf{54.5}  & & 100\% & 22.7\% & 30.6\%\\
    \hline
  \end{tabular}
  \caption{Comparison of required environment steps $n$ for \ac{NP}, \ac{AP}, \ac{PIRL}, and \ac{ACP} with a fixed amount of expert steps $n^{\mathrm{exp}}$ in order to reach $G_{t=0}^{\mathrm{tar}}$. All steps are reported in thousands and the lowest total steps $n^{\mathrm{tot}}$ are bold. The sample reduction gained from \ac{ACP} is reported in percent and computed as $1 - n^{\mathrm{tot}}_{\mathrm{ACP}} / n^{\mathrm{tot}}_{\mathrm{NP|AP|PIRL}}$~. The symbol $-$ indicates that \ac{NP} has not reached $G_{t=0}^{\mathrm{tar}}$ in the given training budget and in this case sample reduction \ac{ACP} vs. \ac{NP} is defined as 100\%.}
  \label{tab:sample_efficiency}
  \vspace*{-0.6cm}
\end{table*}

\subsection{Expert and Rollout Evaluation}

\begin{figure}[t]
  \centering
   \vspace*{0.17cm} 
  \begin{subfigure}{\columnwidth}
    \centering
    \begin{tikzpicture}
\begin{axis}[
  width=\columnwidth,
  height=0.58\columnwidth,
  grid=major,
  xticklabel style={font=\scriptsize},
  yticklabel style={font=\scriptsize},
  every axis plot/.append style={thick},
  legend image post style={mark=none},
  clip=true,
  xlabel={Expert demonstration steps $n^{\mathrm{exp}}$},
  xmode=log,
  log basis x=10,
  ylabel={Normalized return},
  ymin=0, ymax=1.05,
  ytick={0,0.25,0.5, 0.65, 0.85,1},
  yticklabels={$0$,$0.25$,$0.5$, $G_{t=0}^{\mathrm{exp}}$, $0.85$,$G_{t=0}^{\mathrm{tar}}$},
  legend columns=3,
  legend style={
    at={(0.43,1.09)},
    anchor=south,
    font=\scriptsize,
    /tikz/every even column/.append style={column sep=6pt},
    draw=none
  },
  enlarge y limits={upper, value=0.18},
]

\addplot[myblue] coordinates {(0,0) (100, 0.20843355137709846) (200, 0.12018664910180284) (500, 0.7919721686741087) (1000,1)};
\addlegendentry{{\fontsize{6}{8}\selectfont\texttt{Walker2D}}}

\addplot[myblue, dashed] coordinates {(0,0) (100, 0.5329334133580684) (200, 0.3966743421928404) (500, 0.9553825819718038) (1000,1)};
\addlegendentry{{\fontsize{6}{8}\selectfont\texttt{Hopper}}}

\addplot[myblue, dotted] coordinates {(0,0) (10, 0.12939797657490654) (100, 0.6004112045853741) (200, 0.8105004565383309) (500, 0.9555949722069065) (1000,1)};
\addlegendentry{{\fontsize{6}{8}\selectfont\texttt{Ant}}}

\addplot[mygreen] coordinates {(0,0) (10, 0) (20, 0.10845415824287685) (50, 0.8295397727593029) (100, 0.9102284675266016) (127,0.8684094891667941) (247,0.9932576930352086) (363,0.9638533045745378) (451,0.9979453014559987) (589,1) (50000,1)};
\addlegendentry{{\fontsize{6}{8}\selectfont\texttt{Humanoid}}}

\addplot[mygreen, dashed] coordinates {(0,0) (10, 0.4661448653196216) (100,0.69908437687948) (500,0.8973334366434241) (1000,0.9216021822394974) (2000,0.9643389778190624) (3000,0.9715551232206114) (4000,0.9918913052070439) (5000,1)};
\addlegendentry{{\fontsize{6}{8}\selectfont\texttt{Pusher}}}

\addplot[mygreen, dotted] coordinates {(0,0) (50,0) (250,0.4619880673568696) (500,0.6297291672010907) (750,0.650495122840628) (1000,0.6410230151154102) (1250,0.6388085912689567) (1500,0.6591556762196124) (2000,0.6633111486136913) (2500,0.6679991967547686) (50000,0.6679991967547686)};
\addlegendentry{{\fontsize{6}{8}\selectfont\texttt{Reacher}}}

\addplot[mygrey] coordinates {(0,0) (100, 0.42949615320334755) (200, 0.6580347343401486) (500, 0.8829773087742457) (1000, 0.9016561148687013) (2000, 1)};
\addlegendentry{{\fontsize{6}{8}\selectfont\texttt{HalfCheetah}}}

\addplot[mygrey, dashed] coordinates {(0,0) (10, 0.454189976370441) (100, 0.5939890219954059) (200, 0.8884218540514128) (500,1)};
\addlegendentry{{\fontsize{6}{8}\selectfont\texttt{Swimmer}}}

\addplot[mygrey, dotted] coordinates {(0,0) (10, 0.45294133646026197) (100, 0.8148929153478225) (200, 0.8289800136570845) (500, 0.851142887677371) (1000,0.8614272360059346) (2000,0.89998542061525) (3000,0.868536830098711) (4000,1)};
\addlegendentry{{\fontsize{6}{8}\selectfont\texttt{HumanoidStandup}}}

\addplot[myturquoise] coordinates {(0,0) (50,0) (1000,0) (10000,0.0923456790123457) (20000,0.3851851851851852) (25000,0.43703703703703706) (40000, 0.6259259259259259) (50000,0.6814814814814815)};
\addlegendentry{{\fontsize{6}{8}\selectfont\texttt{FetchPush}}}

\addplot[myturquoise, dashed] coordinates {(0,0) (50,0.07939189189189189) (1000,0.07432432432432433) (10000,0.3952702702702703) (20000,0.9611486486486487) (40000,0.9864864864864865) (50000,1)};
\addlegendentry{{\fontsize{6}{8}\selectfont\texttt{FetchPickAndPlace}}}

\addplot[myturquoise, dotted] coordinates {(0,0) (100,0) (1000,0.12916666666666668) (10000,0.29583333333333334) (20000,0.75) (30000,0.5875) (40000,0.5458333333333333) (50000,0.5444444444444443)};
\addlegendentry{{\fontsize{6}{8}\selectfont\texttt{FetchSlide}}}

\addplot[mypink] coordinates {(0,0) (50,0.07181942544459644) (250,0.2024623803009576) (500,0.774281805745554) (1000,0.9726402188782489) (1500,0.9767441860465116) (2000,1)};
\addlegendentry{{\fontsize{6}{8}\selectfont\texttt{FetchReach}}}

\addplot[mypink, dashed] coordinates {(0,0) (296,0.15419999999999998) (1482,1)};
\addlegendentry{{\fontsize{6}{8}\selectfont\texttt{InvertedPendulum}}}

\addplot[mypink, dotted] coordinates {(0,0) (1000,0.0331163091931383) (5000,0.5150322704066638) (10000,1)};
\addlegendentry{{\fontsize{6}{8}\selectfont\texttt{InvertedD.Pendulum}}}

\addplot[myorange, dashed, thick] coordinates {(8,1) (50000,1)};
\addplot[mypink, dashed, thick] coordinates {(8,0.65) (50000,0.65)};

\end{axis}
\end{tikzpicture}
    \vspace{-0.4cm}
    \caption{Environment specific episodic return normalized to $[0,1]$ over the amount of expert demonstration steps with $G_{t=0}^{\mathrm{exp}}=0.65 \cdot G_{t=0}^{\mathrm{tar}}$. }
    \label{fig:ch5_n_expert_evaluation}
  \end{subfigure}

  \vspace{0.3em}

  \begin{subfigure}{\columnwidth}
    \centering
    \begin{tikzpicture}
\begin{axis}[
  width=\columnwidth,
  height=0.58\columnwidth,
  ylabel={Total steps $n^{\mathrm{tot}}$},
  grid=major,
  xticklabel style={font=\scriptsize},
  yticklabel style={font=\scriptsize},
  every axis plot/.append style={thick},
  legend image post style={mark=none},
  ymin=0,
  ymax=130000,
  clip=true,
  xlabel={Rollout steps $n^{\mathrm{rol}}$},
]

\addplot[myblue] coordinates {(0,190607) (1043,109650) (10254,68861) (19846,64453) (29562,66169) (40016,76623) (49487,56094)};
\addplot[only marks, mark=*, draw=myorange, fill=myorange, forget plot] coordinates {(49487,56094)};

\addplot[myblue, dashed] coordinates {(0,87000) (1028,74028) (9910,58910) (19898,30898) (30126,37126) (39954,46954) (49664,56664)};
\addplot[only marks, mark=*, draw=myorange, fill=myorange, forget plot] coordinates {(19898,30898)};

\addplot[myblue, dotted] coordinates {(0,191000) (1747,98747) (9842,106842) (19556,116556) (29270,126270) (40051,109051) (49765,118765)};
\addplot[only marks, mark=*, draw=myorange, fill=myorange, forget plot] coordinates {(1747,98747)};

\addplot[mygreen] coordinates {(0,52087) (979,53066) (9995,60082) (19973,70060) (29983,74070) (40034,66121) (50005,76092)};
\addplot[only marks, mark=*, draw=myorange, fill=myorange, forget plot] coordinates {(0,52087)};

\addplot[mygreen, dashed] coordinates {(0,124100) (1492,111592) (9698,73798) (20142,84242) (29840,93940) (40284,104384) (49982,114082)};
\addplot[only marks, mark=*, draw=myorange, fill=myorange, forget plot] coordinates {(9698,73798)};

\addplot[mygreen, dotted] coordinates {(0,196750) (830,113580) (9960,122710) (19920,58670) (29880,68630) (39840,78590) (49800,88550)};
\addplot[only marks, mark=*, draw=myorange, fill=myorange, forget plot] coordinates {(19920,58670)};

\addplot[mygrey] coordinates {(0,73000) (1605,60605) (9630,62630) (19260,72260) (30495,83495) (40125,93125) (49755,102755)};
\addplot[only marks, mark=*, draw=myorange, fill=myorange, forget plot] coordinates {(1605,60605)};

\addplot[mygrey, dashed] coordinates {(0,55000) (2151,49151) (10755,53755) (19359,60359) (30114,67114) (40869,71869) (49473,80473)};
\addplot[only marks, mark=*, draw=myorange, fill=myorange, forget plot] coordinates {(2151,49151)};

\addplot[mygrey, dotted] coordinates {(0,66000) (2627,68627) (10508,76508) (21016,72016) (28897,79897) (39405,90405) (49913,100913)};
\addplot[only marks, mark=*, draw=myorange, fill=myorange, forget plot] coordinates {(0,66000)};

\addplot[myturquoise] coordinates {(0,175000) (1000,111000) (10000,115000) (20000,115000) (30000,115000) (40000,125000) (50000,130000)};
\addplot[only marks, mark=*, draw=myorange, fill=myorange, forget plot] coordinates {(1000,111000)};

\addplot[myturquoise, dashed] coordinates {(0,210000) (1000,106000) (10000,115000) (20000,120000) (30000,110000) (40000,110000) (50000,120000)};
\addplot[only marks, mark=*, draw=myorange, fill=myorange, forget plot] coordinates {(1000,106000)};

\addplot[myturquoise, dotted] coordinates {(0,110000) (1000,91000) (10000,95000) (20000,105000) (30000,100000) (40000,105000) (50000,115000)};
\addplot[only marks, mark=*, draw=myorange, fill=myorange, forget plot] coordinates {(1000,91000)};

\addplot[mypink] coordinates {(0,70500) (1000,61500) (10000,54500) (20000,56500) (30000,66500) (40000,76500) (50000,86500)};
\addplot[only marks, mark=*, draw=myorange, fill=myorange, forget plot] coordinates {(10000,54500)};

\addplot[mypink, dashed] coordinates {(0,4469) (1026,3063) (2038,3851) (2907,4144) (3976,5213) (5005,6242)};
\addplot[only marks, mark=*, draw=myorange, fill=myorange, forget plot] coordinates {(1026,3063)};

\addplot[mypink, dotted] coordinates {(0,10856) (1030,11886) (2009,12865) (3042,13898) (3978,14834) (4969,15825)};
\addplot[only marks, mark=*, draw=myorange, fill=myorange, forget plot] coordinates {(0,10856)};

\end{axis}
\end{tikzpicture}
    \caption{Sum of demonstration steps, rollout steps and \ac{PPO} steps $n^{\mathrm{tot}}$ over the number of rollout steps $n^{\mathrm{rol}}$. Lowest $n^{\mathrm{tot}}$ is marked orange.}
    \label{fig:ch5_n_rollout_evaluation}
  \end{subfigure}

  \caption{Evaluation of the amount of expert and rollout steps.}
  \label{fig:sample_efficiency_plot}
  \vspace*{-0.6cm}
\end{figure}

The next evaluation analyzes how many expert demonstrations are required to reach the target return $G_{t=0}^{\mathrm{tar}}$. For this experiment, \ac{BC} is applied as described in Sec.~\ref{sec:actor_pretraining}. Across most environments, a clear trend is observed: more demonstration data leads to higher achieved return levels (see  Fig.~\ref{fig:ch5_n_expert_evaluation}). One exception is the \texttt{Reacher} environment, where only about 65\% of
$G_{t=0}^{\mathrm{tar}}$ is reached. Since a non-optimal expert policy is required to validate the proposed approach, we set $c = 0.65$ in Sec.~\ref{sec:sample_efficiency_evaluation}. Overall, these results show that the number of demonstrations required to reach $G_{t=0}^{\mathrm{tar}}$ depends on the environment.

Next, we investigate the impact of the number of rollout steps $n^{\mathrm{rol}}$ used
during \ac{ACP} and investigate whether such rollouts
are required at all.
For each environment, the number of expert steps $n^{\mathrm{exp}}$ is kept
fixed and \ac{ACP} followed by \ac{PPO} fine-tuning is applied, until the target return $G^{\mathrm{tar}}_{t=0}$ is reached. 
The number of rollout steps $n^{\mathrm{rol}}$ is varied, and consequently
the number of \ac{PPO} fine-tuning steps
$n^{\mathrm{fine}}_{\mathrm{PPO}}$ needed to reach
$G^{\mathrm{tar}}_{t=0}$ changes.
Fig.~\ref{fig:ch5_n_rollout_evaluation} shows, for all 15 environments, the total number of
environment steps
$n^{\mathrm{tot}}_{\mathrm{ACP}}$ over the number of rollout steps
$n^{\mathrm{rol}}$.
For each environment, the minimum achieved $n^{\mathrm{tot}}_{\mathrm{ACP}}$ is highlighted.
For 12 out of 15 environments (80.0\%), incorporating rollout data
reduces the total number of required environment steps.
Only three out of the 15 environments (20.0\%) attain their minimum without rollout data at
$n^{\mathrm{rol}} = 0$. Interestingly these are the same environments that do not benefit from \ac{ACP} over \ac{AP}. 
However, across all those environments, a saturation effect can be observed and formulated qualitatively: a
moderate number of rollout steps for critic pretraining improves sample efficiency,
while additional rollout data beyond a certain point yield no
further gains.

\subsection{Extended Step Limit \& Residual Architecture Evaluation}

Finally, the effect of the extended step limit and the residual architecture on the sample efficiency is evaluated. Across all environments, ACP with
extended step limit reduces the required number of environment steps by 10.4\% compared to ACP without extended step limit.
Likewise, ACP with residual architecture (and extended step limit) achieves a reduction of 22.1\% relative to ACP without residual architecture,
empirically showing the effectiveness of these two features.

\section{Conclusion}

In this work, we proposed a method for pretraining the actor and critic networks of \ac{PPO} to improve sample efficiency. By starting from an expert-level policy and value function, \ac{RL} training requires less exploration and experiences reduced distribution shift, leading to faster convergence. Critic pretraining can also mitigate catastrophic forgetting that may occur with actor-only pretraining.
Evaluations across 15 benchmark environments show that actor-critic pretraining reduces the required number of environment steps by an average of 86.1\% compared to no pretraining, 30.9\% compared to actor-only pretraining, and 20.5\% compared to PIRL, a state-of-the-art pretraining approach. In addition, in nine out of 15 environments (60.0\%), \ac{PPO} without pretraining failed to converge to a defined target return. While our experiments focus on robotics and locomotion tasks with continuous action spaces, the proposed approach is also applicable to non-robotic environments and discrete action spaces.

Despite these results, limitations remain. First, expert demonstrations are required, which may not always be available. Second, there is no clear method to determine the required amount of expert or rollout data, as both are environment-specific and non-linear hyperparameters. Finally, in three of 15 tasks (20.0\%), actor--critic pretraining does not improve sample efficiency over actor-only pretraining and the reasons for this remain unclear.

Future work therefore can investigate the conditions under which pretraining is beneficial and develop heuristics for selecting the amount of expert data and rollout data.
Moreover, this work focused exclusively on \ac{PPO} as a representative actor-critic algorithm in a continuous action space.
Further research is required to make adjustments for other actor-critic algorithms, such as Soft Actor-Critic methods. Additional empirical validation in discrete action spaces and real industrial environments would also be valuable.







\section*{ACKNOWLEDGMENT}

This project has received funding from the Ministry of
Science, Research and Arts of the Federal State of Baden-Wuerttemberg within the InnovationsCampus Future Mobility (ICM).
ChatGPT 5.2 was used to assist with language editing and LaTeX formatting.


\renewcommand*{\bibfont}{\small}
\printbibliography

\end{document}